\def\arxivsubmission{1}
\PassOptionsToPackage{table}{xcolor}
\documentclass[10pt,logo]{hitsztechreport}
\usepackage[numbers,sort&compress]{natbib}
\usepackage{multirow,placeins,pgf-pie}
\usepackage{needspace}
\usepackage[noabbrev,nameinlink]{cleveref}
\renewcommand{\today}{November 29, 2026}
\newcommand{\paraheading}[1]{\par\addvspace{6pt}\noindent\textbf{#1.}\enspace}
\newcommand{\topparaheading}[1]{\paraheading{#1}}

\hypersetup{pdftitle={RefocusCraft: Unified Video Refocusing via CoC-Guided State Transitions},pdfauthor={Jialiang Feng, Shengxin Liu, Yong Du, Shengfeng He, Yangyang Xu, Jun Yu},pdfsubject={Technical report},bookmarksnumbered=true}
\newcommand{\paperabstract}{%
Post-capture video refocusing enables flexible control over focus and depth of field. However, image-based methods struggle to maintain temporal consistency when applied frame by frame, while two-stage video pipelines can introduce redundant restoration and propagate errors through an all-in-focus intermediate. We propose RefocusCraft, a unified one-step framework for efficient and temporally coherent video refocusing. Our key idea is to model refocusing as local transitions between source and target defocus states, jointly supporting detail recovery, blur synthesis, and appearance preservation within the same video. Guided by signed source and target circle-of-confusion (CoC) maps, our framework adapts a pretrained video diffusion model to directly generate the refocused video without explicit all-in-focus reconstruction. To align feature interactions with these transitions, we introduce two complementary attention mechanisms. Transition-compatible self-attention aggregates generation features with similar source and target defocus states, while defocus-compatible cross-attention selectively reuses source features to support the desired defocus transitions. Source-conditioned decoding further incorporates fine-grained input features to preserve textures and object boundaries. Experiments on our synthetic and real-world datasets demonstrate strong visual fidelity and temporal coherence across bokeh rendering, defocus deblurring, and refocusing.
}

\usepackage{wrapfig}
\usepackage{float}

\ifdefined\arxivsubmission\else\usepackage{animate}\fi

\makeatletter
\renewcommand\AB@affilsepx{\protect\\\protect\Affilfont}
\makeatother
\renewcommand\Affilfont{\normalfont\fontsize{9}{11}\selectfont}
\title{RefocusCraft: Unified Video Refocusing\\via CoC-Guided State Transitions}
\author[1]{Jialiang Feng}
\author[1]{Shengxin Liu}
\author[2]{Yong Du}
\author[3]{Shengfeng He}
\author[1,\ensuremath{\dagger}]{Yangyang Xu}
\author[1]{Jun Yu}
\affil[1]{Harbin Institute of Technology, Shenzhen}
\affil[2]{Ocean University of China}
\affil[3]{Singapore Management University}
\renewcommand{\reportshorttitle}{RefocusCraft: Unified Video Refocusing}
\paperurl{}
\correspondingauthor{Yangyang Xu}

\makeatletter
\fancypagestyle{firststyle}{%
  \fancyhf{}
  \fancyhead[L]{\includegraphics[height=14mm]{\institutionlogo}}
  \fancyhead[R]{}

}
\let\refocusoriginalmaketitle\maketitle
\renewcommand{\maketitle}{%
  \refocusoriginalmaketitle
  \vspace{-15pt}
  {\noindent\small\textsuperscript{\ensuremath{\dagger}}Corresponding author: Yangyang Xu.\par}
  {\noindent\color{HITBlue}\small\ttfamily\href{https://GenerativeIntelligence-HITSZ.github.io/RefocusCraft/}{https://GenerativeIntelligence-HITSZ.github.io/RefocusCraft/}\par}
  \vspace{7pt}
}
\makeatother

\begin{document}
\maketitle
\begin{figure}[!htbp]
    \centering
    \ifdefined\arxivsubmission
    \href{https://GenerativeIntelligence-HITSZ.github.io/RefocusCraft/}{\includegraphics[page=13,width=\linewidth]{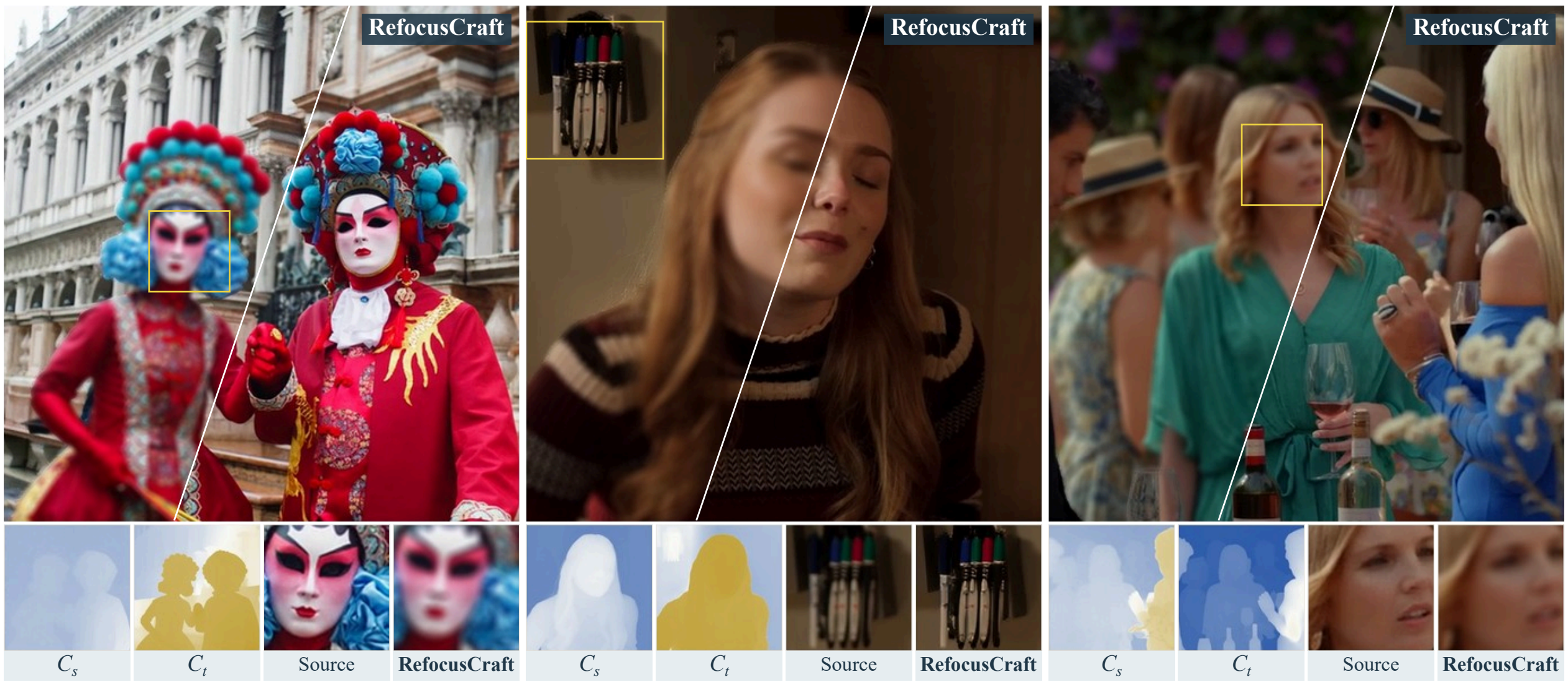}}
    \else
    \animategraphics[type=pdf,width=\linewidth,poster=12,controls,autoplay,loop]{8}{assets/refocus-teaser-animation}{0}{24}
    \fi
    \caption{\textbf{Video refocusing visualization with RefocusCraft.}
    Diagonal wipes reveal our outputs from left to right.
    Signed source/target CoC maps and matched input/output crops are shown
    below each video.
    Blue, white, and yellow denote negative, zero, and positive CoC values,
    respectively. See the \href{https://GenerativeIntelligence-HITSZ.github.io/RefocusCraft/}{Project Page} for video results.}
    \label{fig:real_teaser}
\end{figure}

\par\noindent{\headingfont Abstract}\par\vspace{4pt}
{\small\paperabstract\par}
\clearpage
\section{Introduction}
\label{sec:intro}

\begin{figure}[H]
\centering
\begin{minipage}[t]{.48\linewidth}
\vspace{0pt}

    \centering
    \captionsetup{font=footnotesize,skip=5pt}
    \begin{minipage}[c][56mm][c]{\linewidth}\centering
\includegraphics[width=\linewidth,clip]{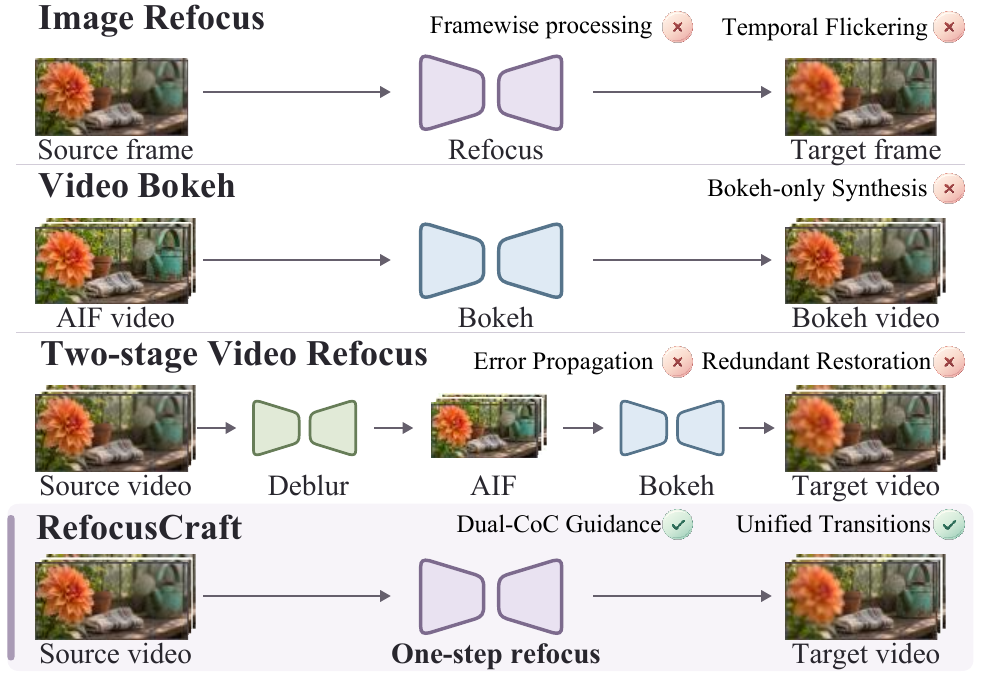}
\end{minipage}

    \captionof{figure}{\textbf{Conceptual comparison of video refocusing pipelines.}
    Framewise image editing, AIF-to-bokeh video rendering, and restore-then-render
    video refocusing are contrasted with RefocusCraft, which directly transforms
    an already-defocused video under dual-CoC guidance without an explicit AIF intermediate.}
    \label{fig:teaser_paradigms}

\end{minipage}\hfill
\begin{minipage}[t]{.48\linewidth}
\vspace{0pt}

    \centering
    \captionsetup{font=footnotesize,skip=5pt}
    \begin{minipage}[c][56mm][c]{\linewidth}\centering
\includegraphics[width=\linewidth,clip]{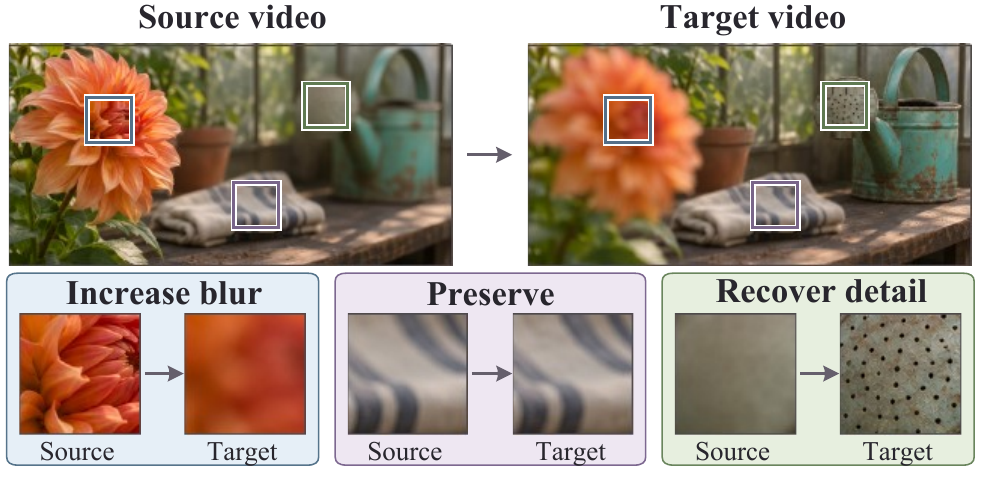}
\end{minipage}

    \captionof{figure}{\textbf{Local source-to-target defocus transitions.}
    Source and target frames illustrate increased blur on the flower (blue),
    appearance preservation on the cloth (purple), and detail recovery on the
    watering-can head (green). The enlarged crops highlight the different conversions
    required within the same scene.}
    \label{fig:teaser_transitions}

\end{minipage}
\end{figure}

Post-capture refocusing enables editors to redirect visual attention and adjust
depth of field in recorded videos without reshooting. 
Refocusing an already-defocused video may require detail recovery, blur synthesis,
and appearance preservation in different regions within the same clip. 
Image-based methods have progressed from separate bokeh
rendering~\cite{peng2022bokehme,zhu2025bokehdiff} and defocus deblurring~\cite{ruan2022learning}
toward unified refocusing. GenFocus reconstructs an all-in-focus
(AIF) image before rendering the desired defocus~\cite{mu2025generative}, whereas DiffCamera
and AnyBokeh directly edit defocused inputs~\cite{wang2025diffcamera,hou2026anybokeh}.
AnyBokeh further uses source and target circle-of-confusion (CoC) maps to describe defocus
changes. 
However, their framewise application to videos does not jointly model temporal
relationships and can produce inconsistent appearance across frames
(first row of Fig.~\ref{fig:teaser_paradigms}).

Video methods primarily address controllable bokeh rendering and photographic
editing~\cite{yang2026any,wang2026bokehcrafter,sun2025generative}, or defocus
deblurring~\cite{morris2025dabit}. 
Existing video bokeh methods primarily synthesize blur from AIF inputs
(second row of Figure~\ref{fig:teaser_paradigms}).
Refocusing Any Video combines restoration and rendering for dynamic scenes by first
reconstructing an AIF video and then synthesizing the desired defocus~\cite{gan2026refocusinganyvideo}
(third row of Figure~\ref{fig:teaser_paradigms}). This restore-then-render pipeline can reconstruct sharp details even in
regions that remain blurred in the target, and intermediate restoration errors can
propagate into subsequent rendering. 
Iterative generative sampling further increases video
processing costs through repeated backbone
evaluations~\cite{mu2025generative,wang2025diffcamera,
hou2026anybokeh,gan2026refocusinganyvideo}.
These limitations motivate a unified framework for
efficient, direct video refocusing.

We formulate video refocusing as local transitions between source and target
defocus states. Figure~\ref{fig:teaser_transitions} illustrates how a single clip can
require increased blur on the flower, preservation of the existing defocus on the cloth,
and detail recovery on the watering-can head. 
Paired source and target CoC maps make
these requirements explicit: the target state specifies the desired defocus, while
the source state determines the conversion required to reach it. 

Following this formulation, we propose RefocusCraft, a unified video refocusing framework
that directly maps source-video latents to target-video latents in a single forward pass,
avoiding iterative sampling and explicit AIF reconstruction
(bottom row of Figure~\ref{fig:teaser_paradigms}). Built on a pretrained video diffusion model~\cite{wan2025wan},
RefocusCraft uses a ControlNet branch~\cite{zhang2023adding} to inject guidance from the
signed source and target CoC maps and their magnitude difference. 
The source latents
provide appearance and motion cues for this transformation.

To align feature interactions with these local transitions, we introduce two complementary
attention mechanisms with CoC-derived spatial visibility masks. Transition-Compatible
Self-Attention (TCSA) permits interactions among generation tokens with similar signed
source and target states, distinguishing regions that share a target blur level but require
different conversions. Defocus-Compatible Cross-Attention (DCCA) then retrieves source
features whose observed CoC magnitudes match the desired target magnitudes, with learned
content similarity determining the aggregation weights. 
Both masks operate on spatial token interactions
within each latent time step, while the retained native attention provides
spatiotemporal context. 
Source-conditioned skip connections between the VAE encoder and decoder
further reuse fine-grained input features to preserve textures and object boundaries. Additionally, we adopt a three-stage progressive training strategy, from single-frame restoration
initialization to unified video refocusing and source-conditioned decoder refinement. Once trained, 
RefocusCraft refocuses source videos in a single forward pass. As shown in Figure~\ref{fig:real_teaser}, RefocusCraft enables flexible
refocusing on real-world videos, recovering details in newly focused
regions and introducing defocus elsewhere to redirect visual attention. Experiments on synthetic and real-world videos demonstrate that our unified framework achieves state-of-the-art performance across bokeh rendering, defocus deblurring, and refocusing.

Our contributions are threefold:
\begin{itemize}[leftmargin=*,nosep]
\item {
We propose RefocusCraft, a unified and efficient framework for one-step video refocusing through
local source-to-target defocus transitions, using dual-CoC conditioning to recover
details, synthesize blur, and preserve appearance without an AIF intermediate.
}

\item {
We design TCSA and DCCA with CoC-derived masks for transition-aware aggregation
and defocus-compatible source retrieval, complemented by source-conditioned
decoding to preserve textures and boundaries.
}

\item {
Quantitative and qualitative evaluations on synthetic and real-world videos
demonstrate state-of-the-art performance across bokeh rendering, defocus
deblurring, and refocusing,
together with efficient video inference.
}
\end{itemize}

\section{Related Work}

\topparaheading{Image Refocus Editing} 
Image-based depth-of-field editing follows three broad strategies. Rendering-based methods synthesize 
defocus from AIF images through depth-guided filtering, layered compositing, or learned 
refinement~\cite{kraus2007depth,wadhwa2018synthetic,peng2022bokehme,Peng2022MPIB,sheng2024dr}. Generative 
methods extend this direction using pretrained priors. BokehDiff~\cite{zhu2025bokehdiff} adopts one-step 
diffusion for lens-blur rendering, while BokehFlow~\cite{huang2026bokehflow} enables depth-free, prompt-controlled 
synthesis through flow matching. Restoration-based refocusing instead recovers a sharp intermediate before 
synthesizing the desired defocus~\cite{bando2007towards,zhang2011single,sakurikar2018refocusgan}. Generative 
Refocusing follows this decomposition with generative deblurring and controllable bokeh synthesis~\cite{mu2025generative}. 
Although this separates restoration from rendering, it can unnecessarily reconstruct regions that remain blurred and 
pass restoration errors to subsequent synthesis. Direct editing methods avoid explicit AIF reconstruction. 
DiffCamera~\cite{wang2025diffcamera} controls refocusing through a target focus point and blur strength, 
while AnyBokeh~\cite{hou2026anybokeh} conditions on signed source and target CoC maps to support spatially varying 
defocus reduction, preservation, and synthesis. These methods establish flexible image refocusing, but their 
framewise application does not jointly model temporal relationships in dynamic scenes.

\paraheading{Video Refocus Editing} 
Video defocus editing additionally requires coherent appearance under scene and camera motion. Earlier bokeh methods 
introduce temporal filtering, feature alignment, and cross-frame aggregation to improve 
consistency~\cite{Zhang2019SyntheticDA,luo2024video}. Recent diffusion-based methods exploit video priors for 
controllable synthesis. Any-to-Bokeh~\cite{yang2026any} incorporates focal-plane geometry, while 
BokehCrafter~\cite{wang2026bokehcrafter} supports text and reference-image guidance. These methods focus on bokeh synthesis from AIF inputs rather than refocusing already-defocused videos. 
CineCtrl~\cite{sun2025generative} extends video diffusion to broader 
photographic effects, including bokeh and exposure. In the restoration direction, video defocus deblurring methods 
recover sharp content without specifying a new target depth of field~\cite{abuolaim2021learning,morris2025dabit}. 
AtlasVDD~\cite{ruan2024self} enables focus correction and bokeh editing through reconstructed sharp neural atlases, 
while RAV~\cite{gan2026refocusinganyvideo} reconstructs an AIF video before rendering the target defocus. RefocusCraft 
instead models source-to-target defocus transitions within a unified video generative framework, jointly supporting 
defocus reduction, increase, and preservation without an explicit AIF-video or sharp-atlas reconstruction stage.

\paraheading{Defocus-Guided Conditioning and Attention} 
Defocus cues guide both generation conditions and feature interactions. Conventional renderers use depth, disparity, 
or CoC maps to determine blur kernels, layered visibility, or neural rendering 
conditions~\cite{peng2022bokehme,Peng2022MPIB,sheng2024dr}. AnyBokeh~\cite{hou2026anybokeh} further uses paired source 
and target CoC maps to describe the required local defocus changes. Such conditioning, however, does not explicitly 
determine which source features are compatible with a desired target defocus level. Attention-based methods impose more 
structured interactions. BokehDiff~\cite{zhu2025bokehdiff} constrains propagation through CoC-defined spatial support and 
occlusion-aware visibility. Any-to-Bokeh~\cite{yang2026any} uses multiplane image masks to structure interactions among 
source and generation tokens. MagicBokeh~\cite{shi2026towards} separates in-focus and out-of-focus interactions through 
target-region masks, while FlashReFocus~\cite{cheng2026flashrefocus} biases attention toward positions with similar 
target defocus. These designs organize propagation according to rendering geometry or target-region relationships. 
RefocusCraft additionally makes source-reference compatibility explicit; DCCA matches source CoC magnitudes to target 
requirements, while TCSA provides complementary generation-side feature aggregation. Their combination coordinates 
source-information reuse and internal feature interaction for video refocusing.

\section{Method}
\label{sec:method}

\subsection{Problem Formulation}
Given a source video $\mathbf{V}_s$ and signed source and target CoC maps
$\mathbf{C}_s$ and $\mathbf{C}_t$, we aim to generate a video
$\hat{\mathbf{V}}_t$ with the desired defocus while preserving scene
content and temporal coherence. CoC magnitude describes blur extent,
while its sign distinguishes the two sides of the corresponding focal
plane. The paired maps specify local source--target transitions:
$\Delta=|\mathbf{C}_t|-|\mathbf{C}_s|$ indicates defocus reduction,
unchanged blur magnitude, or defocus increase through negative, zero,
or positive values, respectively. These requirements can coexist
within the same clip (Fig.~\ref{fig:teaser_transitions}).

\subsection{Overview}
\begin{figure}[!t]
    \centering
    \includegraphics[width=\textwidth,clip]{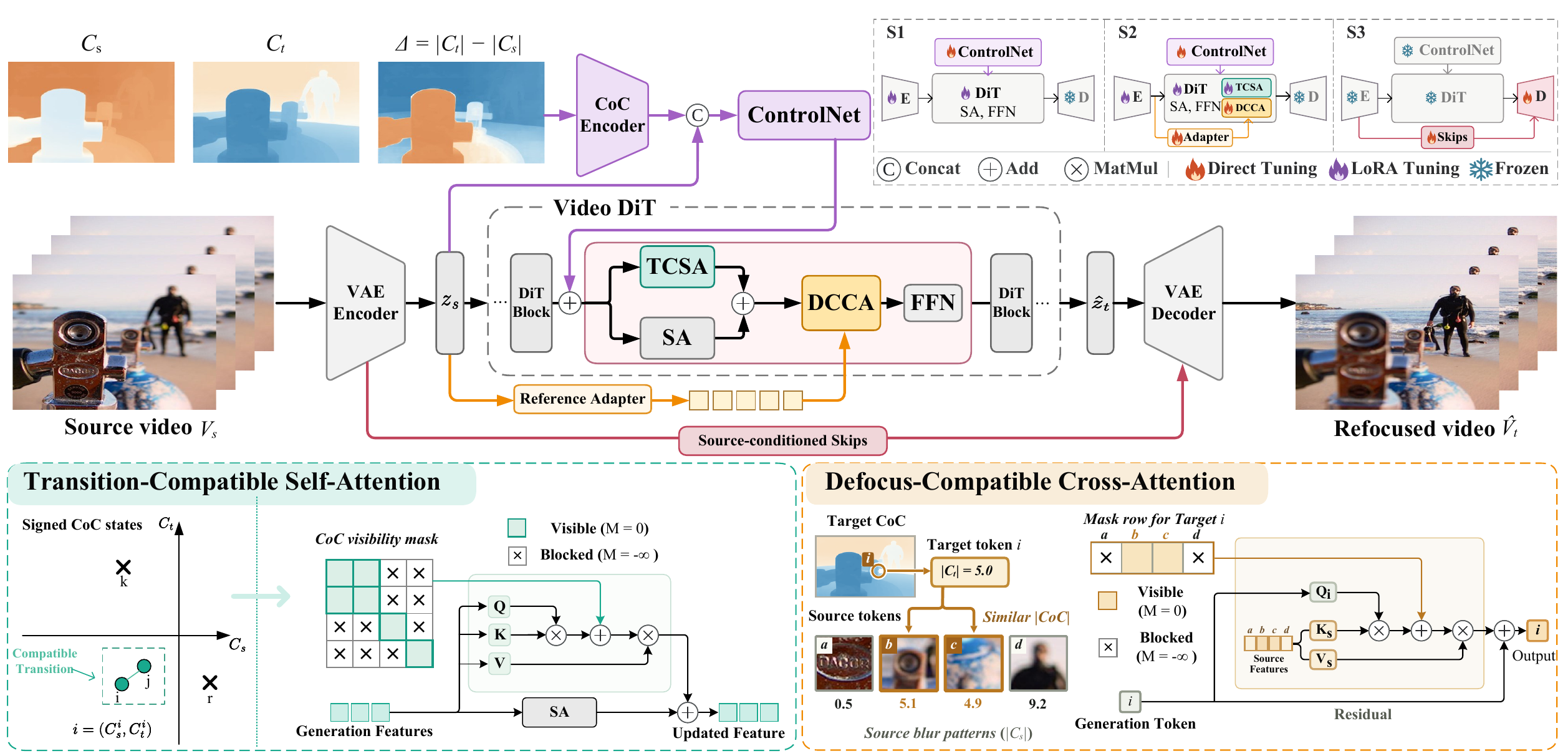}
    \caption{\textbf{Overview of RefocusCraft.}
    \textbf{Top:} The Video DiT predicts target latents in a single forward pass.
    ControlNet conditions the prediction on the source latent, the
    source and target CoC maps, and their magnitude difference.
    Source-conditioned VAE skips help preserve details during decoding.
    \textbf{Lower left (TCSA):} A masked branch runs alongside native
    self-attention and connects generation tokens with similar signed
    source and target CoCs.
    \textbf{Lower right (DCCA):} Generation queries retrieve source
    features whose CoC magnitudes match the desired blur levels.
    Both masks are applied before softmax within each latent time step.
    \textbf{Upper right:} Progressive training comprises image
    deblurring, joint video training, and decoder refinement (Sec.~\ref{sub:progressive_training_strategy}).}
    \label{fig:overview}
\end{figure}

Figure~\ref{fig:overview} illustrates the inference architecture of
RefocusCraft. Our method comprises three parts: (i) a one-step video
refocusing framework with dual-CoC ControlNet guidance and
source-conditioned decoding; (ii) complementary TCSA and DCCA modules
for transition-guided generation-feature interaction and
source-reference retrieval; and (iii) a progressive training strategy
covering spatial restoration, unified video refocusing, and decoder
refinement. The source video is encoded, transformed in latent space,
and decoded without an explicit AIF intermediate.

\noindent\textbf{One-step Video Refocusing.}
Iterative diffusion sampling requires repeated model evaluations and
may introduce source-inconsistent variations~\cite{mu2025generative,
wang2025diffcamera,hou2026anybokeh,wang2026bokehcrafter}.
Since the source video already provides scene appearance and motion,
we predict the target directly from its latent representation,
following recent one-step image and video transformation
methods~\cite{wang2024sinsr,zhu2025bokehdiff,shi2026towards,yang2026any}.

The video VAE encoder $\mathcal{E}_v$ of the pretrained
backbone~\cite{wan2025wan} produces
$\mathbf{z}_s=\mathcal{E}_v(\mathbf{V}_s)$ and retains
multi-scale source features $\mathbf{F}_s$ for decoding.
In parallel, we concatenate the source and target CoC maps
and their magnitude difference along the channel dimension,
then jointly encode this three-channel condition:
\begin{equation}
\mathbf{G}
=
\mathcal{E}_c\!\left(
\operatorname{Concat}(\mathbf{C}_s,\mathbf{C}_t,\Delta)
\right),
\end{equation}
where $\mathcal{E}_c$ denotes the CoC condition encoder.
As shown at the top of Fig.~\ref{fig:overview}, ControlNet~\cite{zhang2023adding}
integrates these conditions with source appearance and injects
conditioning residuals into selected DiT blocks. Starting from
$\mathbf{z}_s$, the adapted network predicts the target latent in one
forward pass, without noise injection or iterative sampling. The
source-conditioned VAE decoder reconstructs the output video:
\begin{equation}
\hat{\mathbf{z}}_t=\mathcal{F}_{\theta}(\mathbf{z}_s;\mathbf{G}),
\quad
\hat{\mathbf{V}}_t=\mathcal{D}_{\psi}(\hat{\mathbf{z}}_t;\mathbf{F}_s).
\end{equation}
Here, $\mathcal{F}_{\theta}$ denotes the complete latent-prediction
network, including the adapted DiT, ControlNet conditioning, and
output head. We adapt the pretrained backbone with
LoRA~\cite{hu2022lora} and jointly optimize the adapters and
conditioning branch. 

We optimize a weighted combination of RGB reconstruction,
perceptual~\cite{zhang2018unreasonable}, and
edge~\cite{zamir2021multi} losses:
\begin{equation}
\mathcal{L}_{\mathrm{pred}}
= \lambda_{\mathrm{rec}}\mathcal{L}_{2}
+ \lambda_{\mathrm{per}}\mathcal{L}_{\mathrm{LPIPS}}
+ \lambda_{\mathrm{edge}}\mathcal{L}_{\mathrm{edge}}.
\end{equation}

\noindent\textbf{Source-conditioned Decoding.}
VAE compression can attenuate fine textures and soften boundaries.
We therefore condition $\mathcal{D}_{\psi}$ on the multi-scale source
features $\mathbf{F}_s$ through the source-conditioned skip
connections~\cite{parmar2024one} in
Fig.~\ref{fig:overview}. At each scale, a separate zero-initialized
$1\times1\times1$ convolution projects the source features, which are
added to the corresponding decoder features before upsampling.
These skips complement the predicted target latent with details preserved in the
input, supporting source-consistent reconstruction of textures and
object boundaries.

\subsection{Defocus Transition-Guided Attention}
ControlNet provides spatial guidance for the required defocus
transitions, but does not explicitly regulate generation-feature
aggregation or source-reference selection. We therefore introduce
TCSA to organize interactions by source--target transition
compatibility, and DCCA to retrieve source evidence compatible with
the target defocus.

\noindent\textbf{Transition-Compatible Self-Attention (TCSA).}
Attention masks in MagicBokeh~\cite{shi2026towards} and
Any-to-Bokeh~\cite{yang2026any} are organized around target focus
regions and focal-plane geometry, respectively, without explicitly
matching the source defocus states of interacting tokens.
For already-defocused inputs, a sharp target may require either
restoring a blurred source or preserving an already-sharp region.
TCSA therefore jointly considers signed source and target CoC values
to distinguish these conversion requirements.

We align the signed CoC maps with the attention-token grid and
construct a spatial mask independently at each latent time step.
In the TCSA branch, a generation query $i$ can attend to a generation
key $j$ only when both their source and target CoCs are close:
\begin{equation}
\mathcal{M}^{\mathrm{T}}_{ij}
=
\begin{cases}
0, &
\max\!\left(|c_i^s-c_j^s|,\,|c_i^t-c_j^t|\right)
\leq\tau_{\mathrm{T}},\\
-\infty, & \text{otherwise},
\end{cases}
\end{equation}
where $c_i^s$ and $c_i^t$ are the aligned source and target CoC values
at token $i$, and $\tau_{\mathrm{T}}$ is the compatibility tolerance.
In the lower-left panel of Fig.~\ref{fig:overview}, $j$ is retained
for query $i$ because both CoC differences satisfy the tolerance;
$k$ and $r$ are excluded because at least one does not. Each such
pairwise decision fills one mask entry. Testing all candidate keys
produces one row, and repeating this process for all queries
constructs the spatial mask.


TCSA runs in parallel with native spatiotemporal self-attention.
Within each latent time step, it applies the transition-compatibility
mask to the attention logits before softmax:
\begin{equation}
\mathbf{A}_{\mathrm{T}}
=
\operatorname{Softmax}\!\left(
\frac{\widetilde{\mathbf{Q}}\widetilde{\mathbf{K}}^{\top}}{\sqrt{d_h}}
+\mathcal{M}^{\mathrm{T}}
\right)\widetilde{\mathbf{V}},
\end{equation}
where $\widetilde{\mathbf{Q}}$, $\widetilde{\mathbf{K}}$, and
$\widetilde{\mathbf{V}}$ are queries, keys, and values computed from
the current generation features, and $d_h$ is the head dimension.
For each query, masked keys receive zero weight after softmax.
The TCSA update thus aggregates features from tokens with compatible
defocus transitions and is added to the native self-attention output.

\noindent\textbf{Defocus-Compatible Cross-Attention (DCCA).}
TCSA guides generation-feature interactions by conversion
compatibility, while DCCA retrieves source evidence compatible with
the target defocus. Since content similarity alone does not ensure
matching blur levels, DCCA filters source references by CoC
magnitude matching.

We align the absolute source and target CoC maps with the token grid,
obtaining source magnitudes $\rho_j^s$ and target magnitudes
$\rho_i^t$. As in TCSA, the mask is constructed independently at each
latent time step. A generation query $i$ can retrieve a source key
$j$ only when their target and source magnitudes are close:
\begin{equation}
\mathcal{M}^{\mathrm{D}}_{ij}
=
\begin{cases}
0, & |\rho_i^t-\rho_j^s|\leq\tau_{\mathrm{D}},\\
-\infty, & \text{otherwise},
\end{cases}
\end{equation}
where $\tau_{\mathrm{D}}$ is the defocus-matching tolerance.
The lower-right panel of Fig.~\ref{fig:overview} illustrates one
mask row: the query has a target magnitude of $5.0$, so source
references $b$ ($5.1$) and $c$ ($4.9$) pass the illustrated tolerance,
while $a$ ($0.5$) and $d$ ($9.2$) are blocked. Repeating this comparison
for each query constructs the generation-to-source mask.

DCCA projects the generation features updated by native
self-attention and TCSA into queries $\mathbf{Q}$. The Reference
Adapter in Fig.~\ref{fig:overview} obtains source keys
$\mathbf{K}_s$ and values $\mathbf{V}_s$ from $\mathbf{z}_s$ through
a shared convolutional projection and block-specific key and value
projections. DCCA aggregates these source features as
\begin{equation}
\mathbf{A}_{\mathrm{D}}
=\operatorname{Softmax}\!\left(
\frac{\mathbf{Q}\mathbf{K}_s^\top}{\sqrt{d_h}}
+\mathcal{M}^{\mathrm{D}}
\right)\mathbf{V}_s.
\end{equation}
The mask selects eligible references, while queries from the
TCSA-refined features determine their aggregation weights.
The result is added back through a separate zero-initialized output
projection, coupling transition-guided generation with source-information
reuse. Queries without a compatible source match receive no
reference update.

\subsection{Progressive Training Strategy}
\label{sub:progressive_training_strategy}
We train RefocusCraft in three stages, progressing from spatial
restoration to unified video refocusing and decoder refinement.

\noindent\textbf{Stage I: Spatial Restoration Initialization.}
Transitions toward sharper targets require recovering details
attenuated by source defocus. We therefore initialize the baseline
through single-frame defocus deblurring with paired AIF targets,
keeping TCSA and DCCA disabled.

\noindent\textbf{Stage II: Unified Spatiotemporal Refocusing.}
Building on this initialization, we introduce TCSA and DCCA and train
on 5-frame clips covering defocus reduction, preservation, and
increase. We then extend training to 13-frame clips to capture
longer temporal dependencies through native spatiotemporal attention.

\noindent\textbf{Stage III: Decoder Refinement.}
To address texture and boundary loss from VAE compression, we freeze
the latent-prediction network and fine-tune the source-conditioned
decoder and its skip projections. This stage refines RGB reconstruction
by learning to incorporate fine-grained source features into the
predicted target appearance.

\section{Experiments}


\subsection{Experimental Setup}

\noindent\textbf{Baselines.}
To assess unified defocus conversion, we compare RefocusCraft
on three tasks: bokeh rendering (AIF-to-Bokeh), defocus
deblurring (Bokeh-to-AIF), and refocusing (Bokeh-to-Bokeh).
Rendering baselines include BokehDiff~\cite{zhu2025bokehdiff} and
Any-to-Bokeh~\cite{yang2026any}.
Refocusing baselines cover AIF-first pipelines,
GenFocus~\cite{mu2025generative} and
RAV~\cite{gan2026refocusinganyvideo}, and direct image editors,
DiffCamera~\cite{wang2025diffcamera} and
AnyBokeh~\cite{hou2026anybokeh}.
We evaluate refocusing methods on all applicable tasks,
applying image-based methods frame by frame.

\noindent\textbf{Datasets.}
Currently, paired real-world videos under different defocus settings
are difficult to obtain. Therefore, we construct a synthetic dataset using scene compositing
and thin-lens-based bokeh rendering following prior
work~\cite{Peng2022MPIB,zhu2025bokehdiff,yang2026any}.
We combine foreground assets from VideoMatte240K and
PhotoMatte85~\cite{lin2021real}, DVM~\cite{sun2021dvm},
and AM-2K~\cite{li2022bridging} with BG-20K and BGMv2 backgrounds~\cite{li2022bridging,lin2021real} to generate
dynamic scenes under multiple defocus configurations.
We pair already-defocused inputs with desired target states
and compute both signed CoC maps from the synthetic disparity
and rendering parameters.
Training uses aligned $512\times512$ crops.
We evaluate on both synthetic and real-world videos.
Using the training synthesis procedure with YouTubeMatte
foregrounds~\cite{yang2025matanyone}, we construct 200 test
scene groups, yielding 400 aligned source-target video pairs
per task. Real-world videos from
UltraVideo~\cite{10.52202/085713-5445} and
CelebV-HQ~\cite{zhu2022celebv} are used for qualitative
comparisons and a user study.
On the synthetic dataset, source CoC is estimated by AnyBokeh~\cite{hou2026anybokeh}. The target CoC follows
the dataset geometry. 

For real-world videos, we estimate source CoC in the same way
and construct target CoC from depth maps predicted by
Video Depth Anything-Large~\cite{chen2025video}, together
with specified focus points and blur strengths.

\noindent\textbf{Evaluation Metrics.}
We report PSNR and SSIM~\cite{wang2004image} for reconstruction
fidelity; LPIPS~\cite{zhang2018unreasonable},
DISTS~\cite{ding2020image}, and FVD~\cite{ge2024content} with VideoMAE-v2 giant
features~\cite{wang2023videomae} for perceptual and video quality.
Following Any-to-Bokeh~\cite{yang2026any}, we use Consistency
(Cons.) and RAFT-based Flow Difference (FD)~\cite{teed2020raft}
to assess temporal consistency and motion fidelity.

\noindent\textbf{Implementation Details.}
We initialize RefocusCraft from Wan2.2-TI2V-5B~\cite{wan2025wan}
and train on aligned $512\times512$ crops with AdamW.
Effective batch sizes are $16$ for single-frame training
and $4$ for video training. Stages~I--II use RGB MSE and
VGG-based LPIPS with weights $1$ and $0.1$.

In Stage~I, we train on single-frame defocus deblurring with
paired AIF targets for 10K steps at $10^{-5}$, with TCSA and
DCCA disabled. Stage~II enables both modules and trains on
5-frame clips across all refocusing tasks for 40K steps,
using learning rates of $10^{-5}$ for existing trainable
modules and $5\times10^{-5}$ for the attention modules.
We then train on 13-frame clips for 10K steps, halving both
learning rates and adding an edge loss weighted by $0.05$.
Stage~III freezes the predictor and fine-tunes the VAE decoder
and its skip projections on 5-frame clips for 30K steps,
at $10^{-5}$ and $10^{-4}$, respectively. It uses RGB $L_1$,
LPIPS, and edge losses weighted by $1$, $0.1$, and $0.1$.
During inference, we process each 25-frame evaluation clip jointly.

\begin{table}[tbp]
\centering
\caption{\textbf{Quantitative comparison on the synthetic dataset across three
defocus-editing tasks.}
The best and second-best results on the full evaluation set
are \textbf{bold} and \underline{underlined}, respectively.}
\label{tab:quantitative_comparison}
\begingroup
\small
\setlength{\tabcolsep}{3pt}
\renewcommand{\arraystretch}{1.15}
\begin{tabular*}{\linewidth}{@{\extracolsep{\fill}}cl*{7}{c}@{}}
\toprule
\textbf{Task} & \textbf{Method}
& \textbf{PSNR}$\uparrow$ & \textbf{SSIM}$\uparrow$
& \textbf{LPIPS}$\downarrow$ & \textbf{DISTS}$\downarrow$
& \textbf{Cons.}$\downarrow$ & \textbf{FD}$\downarrow$
& \textbf{FVD}$\downarrow$ \\
\midrule
\multirow{7}{*}{\rotatebox[origin=c]{90}{\footnotesize\textbf{AIF $\rightarrow$ Bokeh}}} & BokehDiff & 29.00 & 0.8650 & 0.1684 & 0.0848 & 0.0198 & 0.2892 & 37.45 \\
  & AnyBokeh & 26.99 & 0.8998 & 0.1750 & 0.0833 & 0.0190 & 0.3475 & 46.95 \\
  & Any-to-Bokeh & \underline{36.13} & \textbf{0.9770} & \underline{0.0607} & \underline{0.0382} & \textbf{0.0077} & \underline{0.2554} & \underline{19.64} \\
  & GenFocus & 30.09 & 0.9351 & 0.1128 & 0.0572 & 0.0144 & 0.3136 & 22.13 \\
  & RAV & 28.99 & 0.9076 & 0.1278 & 0.0618 & 0.0177 & 0.3254 & 28.94 \\
  & DiffCamera & 25.89 & 0.8537 & 0.2056 & 0.0988 & 0.0192 & 0.2954 & 48.60 \\
  & \textbf{RefocusCraft} & \textbf{36.33} & \underline{0.9762} & \textbf{0.0526} & \textbf{0.0294} & \underline{0.0079} & \textbf{0.1973} & \textbf{15.76} \\
\midrule
\multirow{5}{*}{\rotatebox[origin=c]{90}{\footnotesize\textbf{Bokeh $\rightarrow$ AIF}}}
  & AnyBokeh & 23.00 & 0.7110 & 0.2721 & 0.1170 & 0.0286 & 0.3166 & 75.61 \\
  & GenFocus & \underline{25.14} & \underline{0.7731} & \underline{0.2237} & \underline{0.0895} & 0.0229 & 0.3063 & \underline{38.40} \\
  & RAV & 22.56 & 0.7016 & 0.2805 & 0.1119 & 0.0337 & 0.3615 & 62.23 \\
  & DiffCamera & 24.63 & 0.7583 & 0.2270 & 0.0944 & \underline{0.0227} & \underline{0.2494} & 44.56 \\
  & \textbf{RefocusCraft} & \textbf{27.41} & \textbf{0.8216} & \textbf{0.1785} & \textbf{0.0844} & \textbf{0.0158} & \textbf{0.1952} & \textbf{36.47} \\
\midrule
\multirow{5}{*}{\rotatebox[origin=c]{90}{\footnotesize\textbf{Bokeh $\rightarrow$ Bokeh}}} & AnyBokeh & 26.71 & 0.8753 & 0.1859 & 0.0951 & 0.0204 & 0.3970 & 51.47 \\
  & GenFocus & \underline{28.50} & \underline{0.9024} & \underline{0.1541} & \underline{0.0773} & \underline{0.0180} & 0.4371 & \underline{32.88} \\
  & RAV & 25.78 & 0.8486 & 0.1997 & 0.0950 & 0.0276 & 0.4576 & 53.41 \\
  & DiffCamera & 27.63 & 0.8712 & 0.1924 & 0.0965 & 0.0190 & \underline{0.3546} & 34.68 \\
  & \textbf{RefocusCraft} & \textbf{33.39} & \textbf{0.9381} & \textbf{0.1033} & \textbf{0.0581} & \textbf{0.0113} & \textbf{0.2535} & \textbf{21.12} \\
\bottomrule
\end{tabular*}
\endgroup
\end{table}

\subsection{Results and Comparisons}

\noindent\textbf{Synthetic Videos.}
As shown in Table~\ref{tab:quantitative_comparison}, RefocusCraft
provides strong reconstruction fidelity, perceptual and video
quality, and temporal consistency across three tasks within
a unified framework. On AIF-to-Bokeh, it attains the best PSNR,
LPIPS, DISTS, FD, and FVD among the evaluated methods, and is
second only to Any-to-Bokeh in SSIM and Consistency error,
indicating reconstruction fidelity comparable to the dedicated Any-to-Bokeh model. 
Its PSNR margin over Any-to-Bokeh is only
$0.20$~dB, whereas LPIPS and FVD decrease by $13\%$ and $20\%$,
respectively: the gain thus lies in detail and temporal quality
rather than in global photometric accuracy, consistent with
details being reused from the source rather than regenerated.
On Bokeh-to-AIF, it achieves 27.41 dB PSNR and 0.1785 LPIPS, 
outperforming the strongest baseline by 2.27 dB in PSNR. 
The imperfect AIF reconstructions highlight a potential limitation 
of restore-then-render pipelines: restoration errors may persist 
through subsequent bokeh rendering. This risk motivates our direct 
source-to-target formulation without an explicit AIF intermediate.
The clearest advantage appears on Bokeh-to-Bokeh, where
RefocusCraft outperforms both AIF-first pipelines and framewise
image editors. Its $4.89$~dB PSNR gain over the second-best
result, together with the lowest Consistency error and FD,
indicates improvements in target reconstruction and temporal
agreement.

\begin{figure}[tbp]
    \centering
    \includegraphics[width=\linewidth]{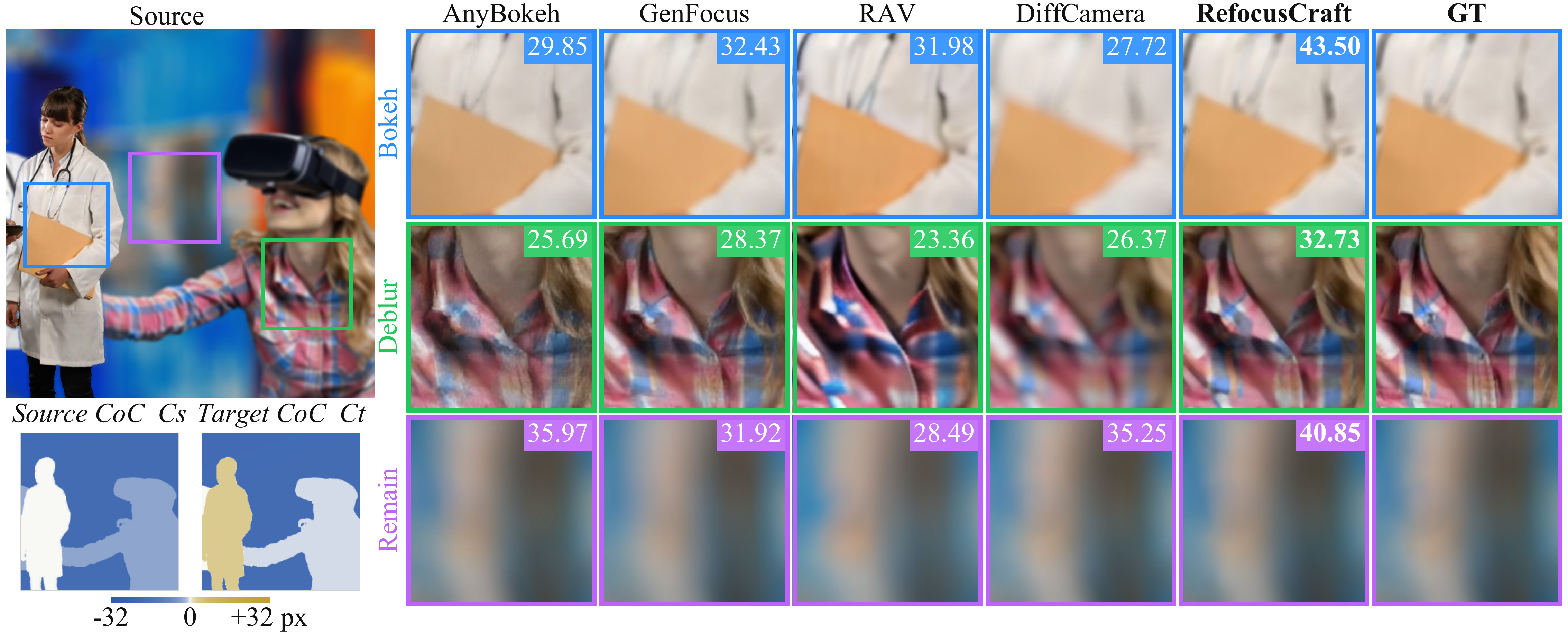}
    \caption{\textbf{Local defocus conversion on synthetic videos.}
    Blue, green, and purple crops illustrate increased blur,
    detail recovery, and defocus preservation, respectively. Top-right numbers report crop PSNR
    (dB). Best viewed zoomed in.}
    \label{fig:synthetic_refocus_comparison}
\end{figure}

Figure~\ref{fig:synthetic_refocus_comparison} illustrates
local blur synthesis, detail recovery, and blur preservation
within a single refocusing example. Competing methods
leave plaid colors mixed during detail recovery or
overemphasize details where stronger blur is required.
RefocusCraft more closely matches the ground-truth textures
and defocus, supporting accurate local conversion within
an already-defocused scene.

\begin{figure}[tbp]
    \centering
    \includegraphics[width=\linewidth]{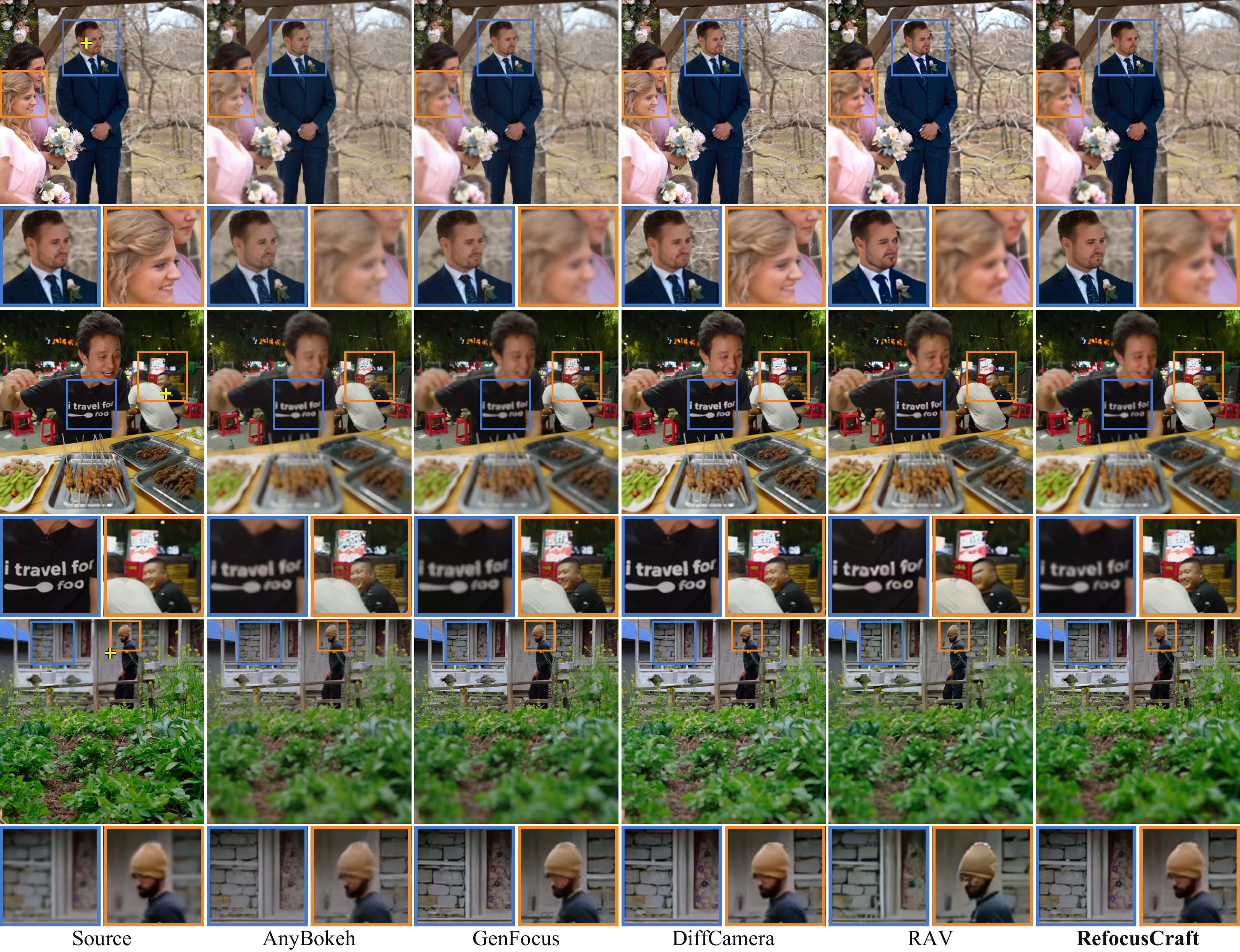}
    \caption{\textbf{Qualitative comparisons on real-world videos.}
    Yellow crosses mark target focus points; colored boxes indicate
    enlarged regions highlighting local details and refocusing effects.}
    \label{fig:real_comparison}
\end{figure}

\noindent\textbf{Real-World Videos.}
Figure~\ref{fig:real_comparison} highlights local defocus control and
content preservation on real-world videos. In the wedding scene,
DiffCamera leaves the foreground face largely sharp and accentuates
background branches, while RAV changes the groom's facial appearance.
RefocusCraft increases foreground blur while keeping the groom's
facial structure, skin tone, and suit details close to the input,
demonstrating selective blur synthesis alongside appearance
preservation. In the garden scene, it makes the stone joints and
window pattern more distinct while softening the foreground foliage.
The subject's hat contour and facial profile also remain close to
the source, illustrating detail recovery with content preservation.
These examples show that RefocusCraft can coordinate detail recovery,
blur synthesis, and appearance preservation within the same scene,
supporting the overall effectiveness of our source-to-target
refocusing framework.

\begin{figure}[tbp]
\centering
\begin{minipage}[t]{.47\linewidth}
\vspace{0pt}\centering
\definecolor{studyBlue}{HTML}{8093BD}
\definecolor{studyOrange}{HTML}{F3BB88}
\definecolor{studyGreen}{HTML}{B1C6A2}
\definecolor{studyRed}{HTML}{E39C92}
\definecolor{studyGray}{HTML}{D3CBC1}
\begin{minipage}[c]{.44\linewidth}
\centering
\tikzset{studyPieNoBorder/.style={draw=none}}
\begin{tikzpicture}
\pie[radius=1.52,rotate=67.68,style=studyPieNoBorder,hide label,hide number,color={studyBlue,studyRed,studyOrange,studyGreen,studyGray}]{65.8/,14.0/,2.8/,8.2/,9.2/}
\end{tikzpicture}
\end{minipage}\hfill
\begin{minipage}[c]{.53\linewidth}
\fontsize{8}{10}\selectfont
\setlength{\tabcolsep}{1.4pt}
\renewcommand{\arraystretch}{1.14}
\begin{tabular}{@{}lr@{}}
\textcolor{studyBlue}{\rule{.5em}{.5em}} RefocusCraft & 65.8\%\\
\textcolor{studyRed}{\rule{.5em}{.5em}} RAV & 14.0\%\\
\textcolor{studyGray}{\rule{.5em}{.5em}} DiffCamera & 9.2\%\\
\textcolor{studyGreen}{\rule{.5em}{.5em}} GenFocus & 8.2\%\\
\textcolor{studyOrange}{\rule{.5em}{.5em}} AnyBokeh & 2.8\%
\end{tabular}
\end{minipage}
\captionof{figure}{\textbf{User preference on the real-world videos.}}
\label{fig:user_study_pie}
\end{minipage}\hfill
\begin{minipage}[t]{.49\linewidth}
\vspace{0pt}\centering
\small
\setlength{\tabcolsep}{5pt}
\renewcommand{\arraystretch}{1.16}
\begin{tabular}{@{}lrr@{}}
\toprule
\textbf{Method} & \textbf{Params}$\downarrow$ (M) & \textbf{Time}$\downarrow$ (s)\\
\midrule
AnyBokeh & 12038.51 & 238.90\\
GenFocus & 12448.84 & 597.65\\
RAV & \textbf{4053.20} & 81.98\\
DiffCamera & 12332.63 & 364.78\\
\midrule
\textbf{RefocusCraft} & 6731.04 & \textbf{2.36}\\
\bottomrule
\end{tabular}
\captionof{table}{\textbf{Inference efficiency for video refocusing.}}
\label{tab:efficiency_comparison}
\end{minipage}
\end{figure}

To assess overall perceptual quality without aligned
real-world ground truth, we conduct an online and in-person
user study with 50 participants on 50 videos from the
real-world collection. Each participant evaluates 20 videos
randomly sampled without replacement and selects the best
result among five methods, considering focus quality,
bokeh naturalness, content fidelity, and temporal coherence.
As shown in Fig.~\ref{fig:user_study_pie}, RefocusCraft
achieves the highest user preference among the compared methods.
These results support the
perceptual quality of our refocusing on real-world videos.
\par

\noindent\textbf{Efficiency.}
We also compare the inference time for processing 25-frame $512\times512$ videos on an NVIDIA A800 GPU with precomputed conditions. 
As shown in Table~\ref{tab:efficiency_comparison},
RefocusCraft achieves the lowest inference time
($2.36$~s per clip) and uses fewer parameters ($6.73$B)
than the image-based baselines.
AnyBokeh and DiffCamera perform iterative editing frame
by frame, while GenFocus and RAV restore an AIF intermediate
before rendering.
Our direct one-step video mapping avoids iterative sampling
and explicit AIF reconstruction, reducing repeated
backbone evaluations.

\subsection{Ablation Studies}
We ablate components on Bokeh-to-Bokeh (Table~\ref{tab:component_ablation}). Baseline removes both TCSA and DCCA. 
w/o TCSA and w/o DCCA remove the corresponding attention branch, whereas w/o SD pairs the full latent predictor 
with the original frozen VAE decoder. 

\begin{table}[tbp]
\centering
\caption{\textbf{Component ablation on Bokeh-to-Bokeh.}}
\label{tab:component_ablation}
\begingroup
\small
\setlength{\tabcolsep}{7pt}
\renewcommand{\arraystretch}{1.18}
\begin{tabular}{@{}l*{6}{c}@{}}
\toprule
\textbf{Variant}
& \textbf{PSNR}$\uparrow$ & \textbf{SSIM}$\uparrow$ & \textbf{LPIPS}$\downarrow$
& \textbf{Cons.}$\downarrow$ & \textbf{FD}$\downarrow$ & \textbf{FVD}$\downarrow$ \\
\midrule
Baseline  & 31.28 & 0.9126 & 0.1324 & 0.0161 & 0.3658 & 30.76 \\
w/o TCSA  & 33.33 & 0.9375 & 0.1048 & 0.0114 & 0.2565 & 21.39 \\
w/o DCCA  & 33.14 & 0.9364 & 0.1055 & 0.0117 & 0.2561 & 21.77 \\
w/o SD    & 30.44 & 0.8967 & 0.1425 & 0.0188 & 0.4319 & 33.89 \\
\midrule
\textbf{Full model}
& \textbf{33.39} & \textbf{0.9381} & \textbf{0.1033} & \textbf{0.0113} & \textbf{0.2535} & \textbf{21.12} \\
\bottomrule
\end{tabular}
\endgroup
\end{table}

Adding either TCSA or DCCA improves all reported metrics
over Baseline. Combining them further improves all metrics,
including PSNR gains of $0.06$~dB over DCCA alone and
$0.25$~dB over TCSA alone, supporting complementary benefits
from generation-feature interaction and source-reference retrieval.
Relative to Baseline, the full model improves PSNR by
$2.11$~dB and reduces LPIPS and FD by $22\%$ and $30.7\%$,
respectively.
Compared with w/o SD, it increases PSNR from $30.44$ to
$33.39$~dB and reduces FVD from $33.89$ to $21.12$,
supporting the combined benefit of source-conditioned
decoding and decoder refinement. This decoding configuration
reuses fine-grained source features to mitigate texture and
boundary loss from VAE compression.Together, these results 
support combining defocus-guided
attention with source-conditioned decoding for accurate
and temporally coherent video refocusing.
The supplementary material further separates decoder fine-tuning
from source skip connections and provides additional visual comparisons. 
\section{Conclusion}
In this work, we present RefocusCraft, a unified framework for efficient
one-step video refocusing. Guided by signed source and
target CoC maps, it directly models local defocus transitions
without iterative sampling or an explicit all-in-focus
intermediate.
TCSA and DCCA provide mutually reinforcing feature guidance,
complemented by progressive training and source-conditioned
decoding for detail recovery and preservation.
Experiments on synthetic and real-world videos demonstrate
effective bokeh rendering, defocus deblurring, and refocusing,
combining target-consistent defocus control with content
fidelity and temporal coherence.
These capabilities offer potential for focus correction
and creative depth-of-field editing in video post-production.

\FloatBarrier
{\small
\setlength{\parskip}{0pt}
\setlength{\bibsep}{0pt plus .3pt}
\bibliographystyle{ieeenat_fullname}
\bibliography{main}
}
\clearpage
\appendix
\linespread{1.0}\selectfont
\setlength{\parskip}{3pt plus .5pt minus .5pt}
\setlength{\textfloatsep}{9pt plus 2pt minus 1pt}
\setlength{\floatsep}{8pt plus 2pt minus 1pt}
\setlength{\intextsep}{8pt plus 2pt minus 1pt}
\captionsetup{skip=5pt}
\setcounter{figure}{0}
\setcounter{table}{0}
\setcounter{equation}{0}
\renewcommand{\thefigure}{S\arabic{figure}}
\renewcommand{\thetable}{S\arabic{table}}
\renewcommand{\theequation}{S\arabic{equation}}
\renewcommand{\theHfigure}{supp.\arabic{figure}}
\renewcommand{\theHtable}{supp.\arabic{table}}
\renewcommand{\theHequation}{supp.\arabic{equation}}
{\color{HITBlue}\titlefont Supplementary Material\par}
\medskip
Video results are available on the \href{https://GenerativeIntelligence-HITSZ.github.io/RefocusCraft/}{Project Page}.\par
\medskip
\noindent\begin{minipage}{\textwidth}
    \captionsetup{type=figure}
    \centering
    \includegraphics[width=\linewidth]{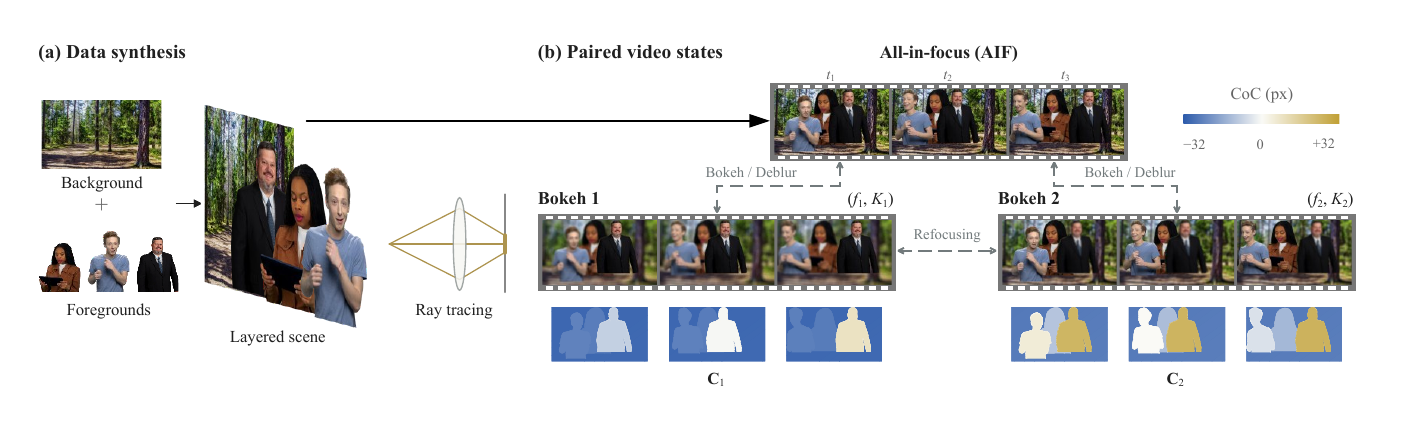}
    \caption{\textbf{Synthetic dataset construction and paired supervision.}
    (a) Foreground assets and a background are composited into a dynamic scene,
    followed by AIF construction and thin-lens bokeh rendering.
    (b) The AIF video and two bokeh videos form six cross-state directed pairs
    for bokeh rendering, defocus deblurring, and refocusing.
    Corresponding signed CoC maps are shown below the bokeh frames.
    Solid arrows indicate synthesis; dashed arrows indicate paired supervision.}
    \label{fig:supp_dataset_pipeline}
\end{minipage}\par

\section{Synthetic Dataset Construction}
\label{sec:supp_dataset}

Following MPIB~\cite{Peng2022MPIB},
BokehDiff~\cite{zhu2025bokehdiff}, and
Any-to-Bokeh~\cite{yang2026any}, we construct a synthetic video
dataset through scene compositing and layered thin-lens rendering.
Foreground assets are collected from VideoMatte240K and
PhotoMatte85~\cite{lin2021real}, DVM~\cite{sun2021dvm},
and AM-2K~\cite{li2022bridging}, while backgrounds are drawn
from BG-20K and BGMv2~\cite{li2022bridging,lin2021real}.

For each scene, we randomly select a background and two video
foregrounds, optionally adding image foregrounds to increase
scene diversity. Foregrounds are resized according to their
visible alpha regions, with randomly sampled initial positions
and disparities. During the sequence, foreground positions
follow random-walk trajectories, while their scales and
disparities can vary smoothly to simulate approaching or
receding motion. Video foregrounds also retain their original
appearance changes and motion. Alpha compositing produces
an AIF sequence together with its aligned disparity maps.

We generate two bokeh sequences by separately sampling their
focus and blur settings. For each sequence, we sample a blur
strength $K_i \sim \mathcal{U}(0,32)$ and randomly select a
foreground track whose median disparity determines the focus
disparity $f_i$. Rendering uses a fronto-parallel layered
approximation, with representative layer disparities obtained
from the scene metadata. For each output pixel, we trace rays
through different locations within a circular aperture and
sample the scene layers at their intersections. Layer
contributions are composited according to visibility and
aggregated across aperture samples.

Each scene group contains three aligned video states---AIF 
and two bokeh sequences---providing six cross-state directed
pairs for bokeh rendering, defocus deblurring, and refocusing.
Signed CoC annotations are computed from the saved disparity
maps and rendering parameters:
\begin{equation}
    \mathbf{C}_{i,\tau}
    = K_i \bigl(\mathbf{D}_{\tau} - f_i\bigr),
    \qquad
    \mathbf{C}_{\mathrm{AIF},\tau} = 0,
    \label{eq:supp_synthetic_coc}
\end{equation}
where $\tau$ denotes the frame index and $i \in \{1,2\}$
indexes the bokeh state. Each source--target pair is associated
with its corresponding source and target CoC maps. Same-state
pairs are additionally retained for identity supervision.

Using 1,300 training backgrounds, we generate 19,500 scene
groups. Each video contains 25 frames at a resolution of
$1024 \times 576$. For evaluation, we apply the same synthesis
procedure with YouTubeMatte foregrounds~\cite{yang2025matanyone}
and 200 backgrounds disjoint from training, yielding 200 scene
groups and 400 video pairs per task. The synthesis pipeline
and representative examples are shown in
Fig.~\ref{fig:supp_dataset_pipeline}.

\Needspace{.56\textheight}
\section{User Study Details}
\label{sec:supp_user_study}

\begin{wrapfigure}{r}{.43\textwidth}
    \centering
    \includegraphics[width=\linewidth]{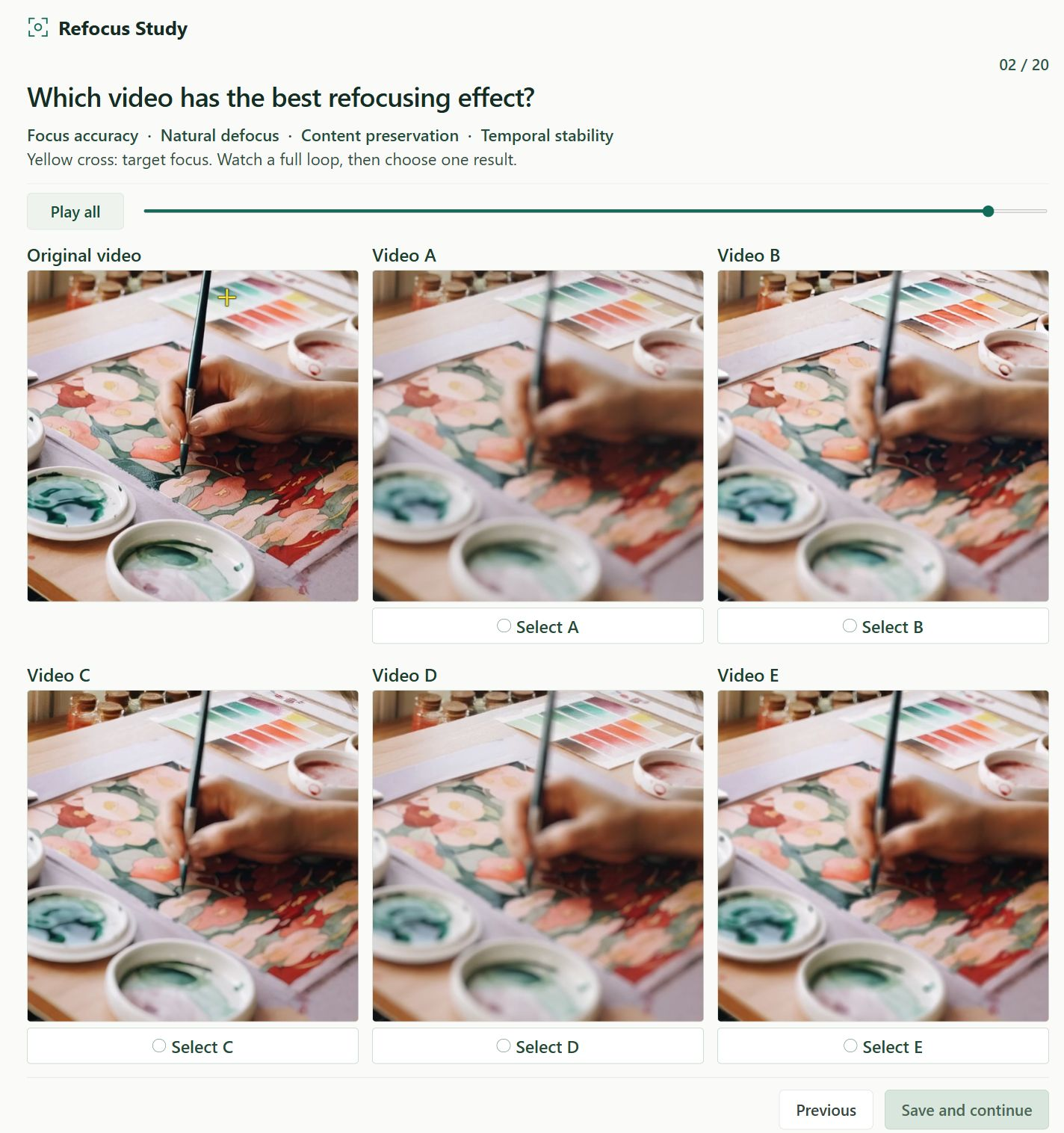}
    \caption{\textbf{User study interface.}
    Participants compare the source video with five anonymized candidates
    and select the best overall refocusing result. The yellow cross marks
    the target focus point. Candidate order is randomized, and the interface
    supports synchronized playback and enlarged viewing.}
    \label{fig:supp_user_study_interface}
\end{wrapfigure}

We conduct a user study to evaluate perceptual refocusing quality
on real-world videos without aligned ground-truth targets.
We compare RefocusCraft with AnyBokeh, GenFocus, DiffCamera,
and RAV. All methods process the same source frames under a
shared target focus and defocus configuration. Each clip contains
25 frames at $512\times512$ resolution. Each participant evaluates
20 distinct videos randomly sampled without replacement from
the real-world collection.

Figure~\ref{fig:supp_user_study_interface} shows a simplified
view of the study interface. Each trial presents the source video
alongside five anonymized results, with a yellow cross marking
the desired focus point in the source. Candidate order is
independently randomized for each trial and participant to
reduce positional bias. Participants watch the videos through
synchronized playback for at least one complete loop before
making a selection, and may replay or enlarge them for inspection.

Participants select the single best overall result by considering
four criteria: appropriate sharpness around the target focus
point, natural out-of-focus blur, preservation of scene content
and motion, and temporal stability without additional flicker,
texture changes, or edge jitter. These criteria guide one overall
choice per trial. We report each method's preference share as
its number of selections divided by the total number of valid
selections.

\par\FloatBarrier

\section{Additional Ablation Studies}
\label{sec:supp_ablation}

\subsection{Training Strategy}
\label{sec:supp_training_ablation}
We ablate the progressive training schedule to evaluate its
contribution to video refocusing.
All variants use 60K optimizer updates, disable TCSA and DCCA,
and retain the original frozen VAE decoder.
\textit{5F only} trains directly on 5-frame clips for 60K updates.
\textit{w/o 13F} performs 10K updates of single-frame defocus
deblurring initialization, followed by 50K updates on 5-frame clips.
\textit{Full schedule} uses the same initialization, followed by
40K updates on 5-frame clips and 10K updates on 13-frame clips
with the corresponding continuation settings.

\begin{table}[htbp]
\centering
\caption{\textbf{Training strategy ablation on video refocusing.}
All variants disable TCSA and DCCA and use the original frozen VAE decoder.
Each schedule totals 60K optimizer updates. Evaluation uses 200 synthetic
scene groups (400 Bokeh-to-Bokeh video pairs), with joint 25-frame inference
on fixed $512\times512$ crops.}
\label{tab:training_ablation}
\begingroup
\fontsize{9}{11}\selectfont
\setlength{\tabcolsep}{1.8pt}
\renewcommand{\arraystretch}{1.16}
\begin{tabular*}{\linewidth}{@{\extracolsep{\fill}}lcc*{7}{c}@{}}
\toprule
Variant & Init. & 13F
& PSNR$\uparrow$ & SSIM$\uparrow$ & LPIPS-V$\downarrow$
& DISTS$\downarrow$ & Cons.$\downarrow$ & FD$\downarrow$
& FVD$\downarrow$ \\
\midrule
5F only & $\times$ & $\times$ & 29.21 & 0.8855 & 0.1678 & 0.08066 & 0.02078 & 0.4969 & 37.52 \\
w/o 13F & $\checkmark$ & $\times$ & 28.86 & 0.8834 & 0.1656 & 0.08052 & 0.02104 & 0.4880 & 38.98 \\
\textbf{Full schedule} & $\checkmark$ & $\checkmark$ & \textbf{30.27} & \textbf{0.8960} & \textbf{0.1443} & \textbf{0.07261} & \textbf{0.01892} & \textbf{0.4350} & \textbf{34.35} \\

\bottomrule
\end{tabular*}

\endgroup
\end{table}

Although \textit{w/o 13F} does not improve every metric
over \textit{5F only} in Table~\ref{tab:training_ablation},
it achieves lower LPIPS-V, DISTS, and FD, suggesting
modest benefits from single-frame initialization in
perceptual similarity and motion fidelity.

\textit{Full schedule} achieves the best results across
all seven metrics. It reaches $30.27$~dB PSNR,
exceeding \textit{5F only} and \textit{w/o 13F}
by $1.06$~dB and $1.41$~dB, respectively.
Compared with \textit{w/o 13F}, LPIPS-V decreases
from $0.1656$ to $0.1443$, while FVD decreases
from $38.98$ to $34.35$, corresponding to reductions
of $12.9\%$ and $11.9\%$, respectively.
Lower Cons.\ and FD further indicate improved temporal
consistency and motion fidelity.
These consistent improvements under joint 25-frame
inference support the effectiveness of the complete
progressive training recipe in improving reconstruction
fidelity, perceptual quality, and temporal performance.

\subsection{Decoder Fine-tuning and Source Skips}
\label{sec:supp_decoder_ablation}
We ablate decoder fine-tuning and source feature reuse to
evaluate their contributions to RGB reconstruction.
All configurations share the same frozen latent predictor.
\textit{Frozen decoder} uses the original frozen VAE decoder;
\textit{w/o source skips} fine-tunes the decoder without source
skip connections; and \textit{Full model} jointly fine-tunes
the decoder and source skip projections.
The two fine-tuned variants follow the same 30K-step
decoder-stage training schedule.

As shown in Table~\ref{tab:supp_decoder_ablation}, decoder
fine-tuning alone slightly improves SSIM and LPIPS-V,
while PSNR remains at $30.44$~dB and FVD increases from
$33.89$ to $39.13$. Decoder adaptation alone thus yields
limited and inconsistent gains.

Adding source skips improves all seven metrics over
decoder-only fine-tuning, and the full model achieves the
best results among the three configurations.
Compared with \textit{w/o source skips}, it increases PSNR
by $2.95$~dB and reduces LPIPS-V and FVD by $25.9\%$ and
$46.0\%$, respectively. Temporal consistency and motion
fidelity also improve, with FD decreasing from $0.4397$
to $0.2535$.
\begin{table}[htbp]
\centering
\caption{\textbf{Decoder fine-tuning and source skips on video refocusing.}
All variants use the same latent predictor, frozen after 60K updates.
Frozen decoder uses the original frozen VAE decoder; w/o source skips
fine-tunes only the decoder; Full model fine-tunes the decoder and source skip projections.
Both fine-tuned variants use 30K decoder-stage updates.
Evaluation uses 200 synthetic scene groups (400 Bokeh-to-Bokeh video pairs),
with joint 25-frame inference on fixed $512\times512$ crops.}
\label{tab:supp_decoder_ablation}
\begingroup
\fontsize{9}{11}\selectfont
\setlength{\tabcolsep}{1.8pt}
\renewcommand{\arraystretch}{1.16}
\begin{tabular*}{\linewidth}{@{\extracolsep{\fill}}lcc*{7}{c}@{}}
\toprule
Variant & FT & Skip
& PSNR$\uparrow$ & SSIM$\uparrow$ & LPIPS-V$\downarrow$
& DISTS$\downarrow$ & Cons.$\downarrow$ & FD$\downarrow$
& FVD$\downarrow$ \\
\midrule
Frozen decoder & $\times$ & $\times$ & 30.44 & 0.8967 & 0.1425 & 0.07127 & 0.01877 & 0.4319 & 33.89 \\
w/o source skips & $\checkmark$ & $\times$ & 30.44 & 0.9014 & 0.1395 & 0.07180 & 0.01911 & 0.4397 & 39.13 \\
\midrule
\textbf{Full model} & $\checkmark$ & $\checkmark$ & \textbf{33.39} & \textbf{0.9381} & \textbf{0.1033} & \textbf{0.05809} & \textbf{0.01133} & \textbf{0.2535} & \textbf{21.12} \\
\bottomrule
\end{tabular*}

\endgroup
\end{table}

Since the latent predictor is fixed, these improvements
highlight the contribution of source feature reuse during
decoding. The results support our source-conditioned
decoding design for improving reconstruction fidelity,
perceptual quality, and temporal consistency.

\section{CoC Robustness}
We evaluate robustness to source CoC estimation errors by
comparing ground-truth source CoCs with estimates from
AnyBokeh~\cite{hou2026anybokeh}, while fixing the model
weights, source video, and target CoC.
Evaluation uses 200 synthetic scene groups, comprising
400 Bokeh-to-Bokeh video pairs.
As shown in Table~\ref{tab:supp_coc_refocus}, RefocusCraft
achieves $33.39$~dB PSNR and $0.9381$ SSIM with estimated
source CoCs, compared with $35.02$~dB PSNR using ground-truth
source CoCs. Although accurate conditioning improves
reconstruction fidelity, the model retains effective
refocusing performance with predicted source conditions.

\begin{figure}[htbp]
    
    \centering
    \begin{minipage}[c]{.55\linewidth}
    \includegraphics[width=\linewidth]{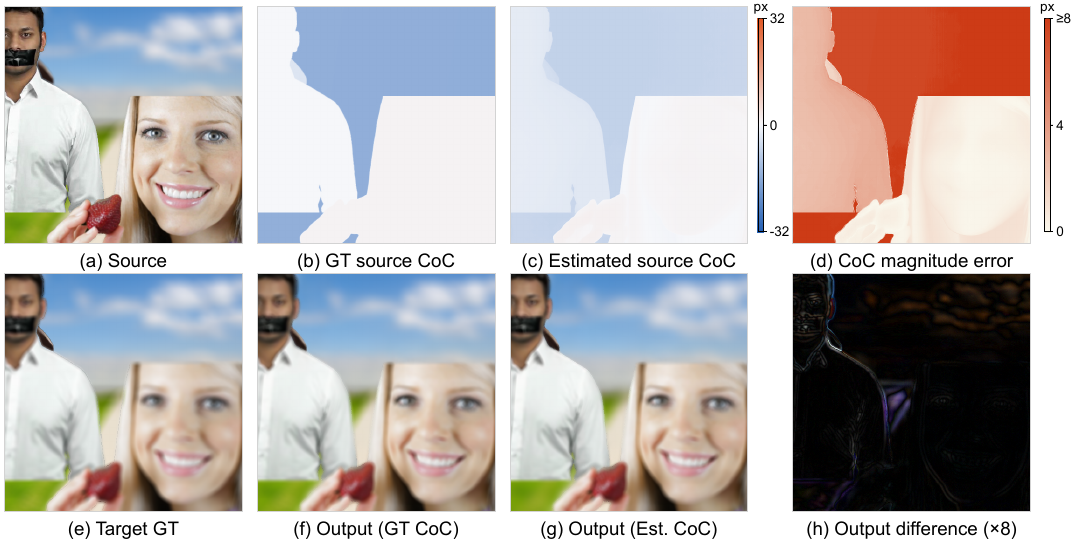}
    \end{minipage}\hfill
    \begin{minipage}[c]{.42\linewidth}
    \caption{\textbf{Qualitative robustness to source CoC estimation errors.}
    (a) Source frame; (b,c) ground-truth and estimated source CoCs;
    (d) absolute difference between their magnitudes;
    (e) target ground truth; (f,g) outputs conditioned on (b,c);
    (h) absolute RGB difference between (f,g), amplified by $8\times$
    for visualization. The model, source video, and target CoC are fixed,
    and all panels show the same frame. The source CoCs share a signed-radius
    color scale; the magnitude-error scale in (d) saturates at $8$ pixels.}
    \label{fig:supp_coc_robustness}
    \end{minipage}
\end{figure}

Figure~\ref{fig:supp_coc_robustness} provides a qualitative
example of this tolerance. Despite noticeable differences
between the ground-truth and estimated source CoC maps,
the two outputs exhibit similar scene appearance and
defocus distributions. The amplified RGB difference map
reveals residual differences in local details, while the
overall refocusing effect remains similar.
Together, these results support the robustness of
RefocusCraft to the source CoC estimation errors evaluated
here and its practical use with predicted CoC maps.

\par
\begin{table}[htbp]
\centering
\caption{\textbf{Sensitivity to source CoC estimation on video refocusing.}
Evaluation uses 200 synthetic scene groups (400 Bokeh-to-Bokeh video pairs;
25 frames per video). The 90K checkpoint, source video,
and target CoC are fixed at inference. Estimated source CoCs are obtained
from AnyBokeh. $\Delta$ is the metric change from GT to estimated source CoC.}
\label{tab:supp_coc_refocus}
\begingroup
\fontsize{9}{11}\selectfont
\setlength{\tabcolsep}{3.2pt}
\renewcommand{\arraystretch}{1.16}
\begin{tabular*}{\linewidth}{@{\extracolsep{\fill}}l*{7}{r}@{}}
\toprule
Source CoC & PSNR$\uparrow$ & SSIM$\uparrow$ & LPIPS-V$\downarrow$ & DISTS$\downarrow$ & Cons.$\downarrow$ & FD$\downarrow$ & FVD$\downarrow$ \\
\midrule
GT & 35.02 & 0.9525 & 0.07716 & 0.03976 & 0.01027 & 0.2189 & 15.19 \\
Estimated & 33.39 & 0.9381 & 0.1033 & 0.05809 & 0.01133 & 0.2535 & 21.12 \\
$\Delta$ & -1.633 & -0.01438 & +0.02617 & +0.01833 & +0.00106 & +0.03463 & +5.928 \\
\bottomrule
\end{tabular*}

\endgroup
\end{table}

\section{More Visual Results}
\label{sec:supp_visual_results}

\subsection{Temporal Consistency}
\begin{figure}[htbp]
    
    \centering
    \includegraphics[width=\linewidth]{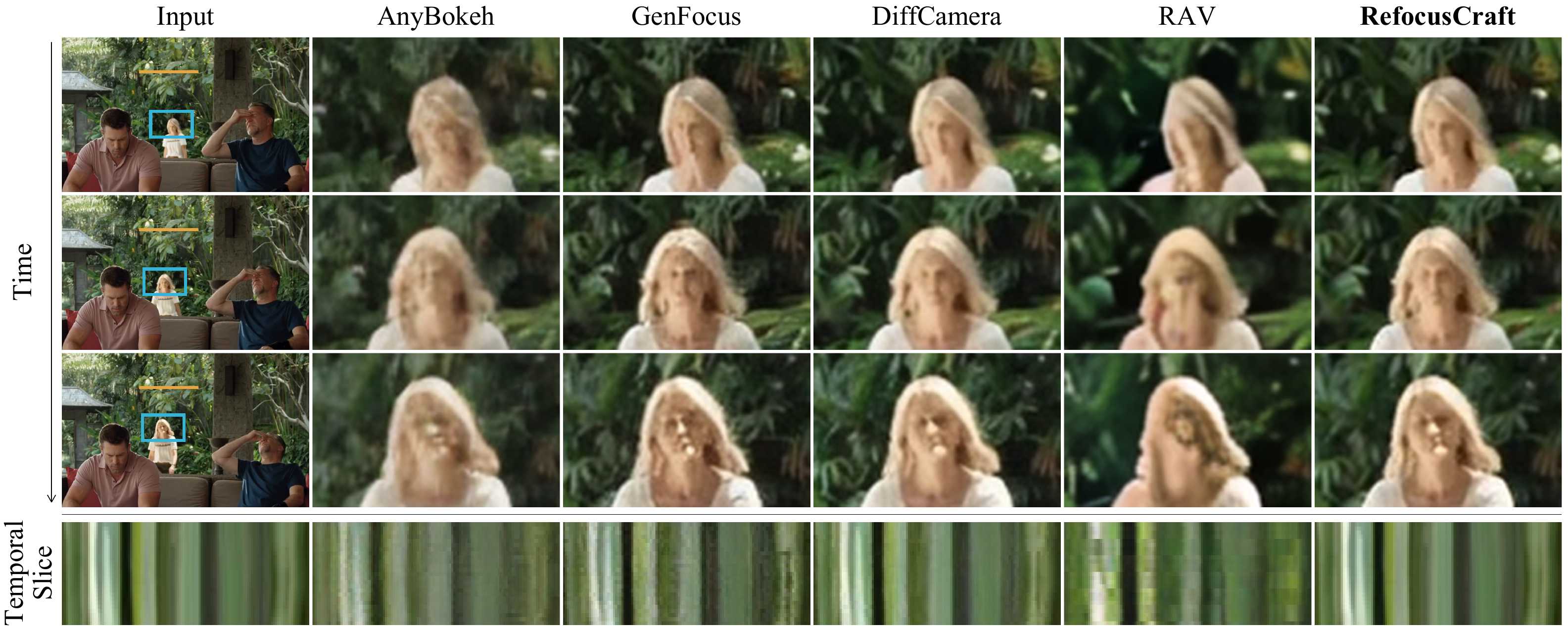}
    \caption{\textbf{Temporal consistency on a real-world video.}
    Cyan boxes indicate the facial regions enlarged in the result columns;
    orange lines mark the background scanlines used for the bottom temporal
    slices. Each slice stacks the same scanline over 25 frames, with time
    increasing downward.}
    \label{fig:real_temporal_comparison}
\end{figure}

\label{sec:supp_temporal}

Figure~\ref{fig:real_temporal_comparison} shows enlarged
facial regions at three time steps. AnyBokeh leaves facial
details visibly smoothed, while RAV introduces noticeable
changes in facial structure and appearance.
RefocusCraft retains more distinct facial contours than
AnyBokeh and maintains a more consistent appearance across
the sampled frames, supporting content preservation as
the subject moves.

The bottom temporal slices stack the same background
scanline over 25 frames, with time increasing downward.
GenFocus exhibits dense horizontal fluctuations, while
DiffCamera and RAV show local banding and abrupt changes
in color and intensity. RefocusCraft produces more
continuous bands along the temporal axis, with fewer
abrupt row-to-row variations, indicating more stable
background appearance. Together, the facial comparisons
and temporal slices illustrate the effectiveness of
RefocusCraft in preserving scene content and maintaining
temporal coherence during video refocusing.

\subsection{Focus and Blur Strength Control}
\begin{figure}[htbp]
    \centering
    \includegraphics[width=.90\linewidth]{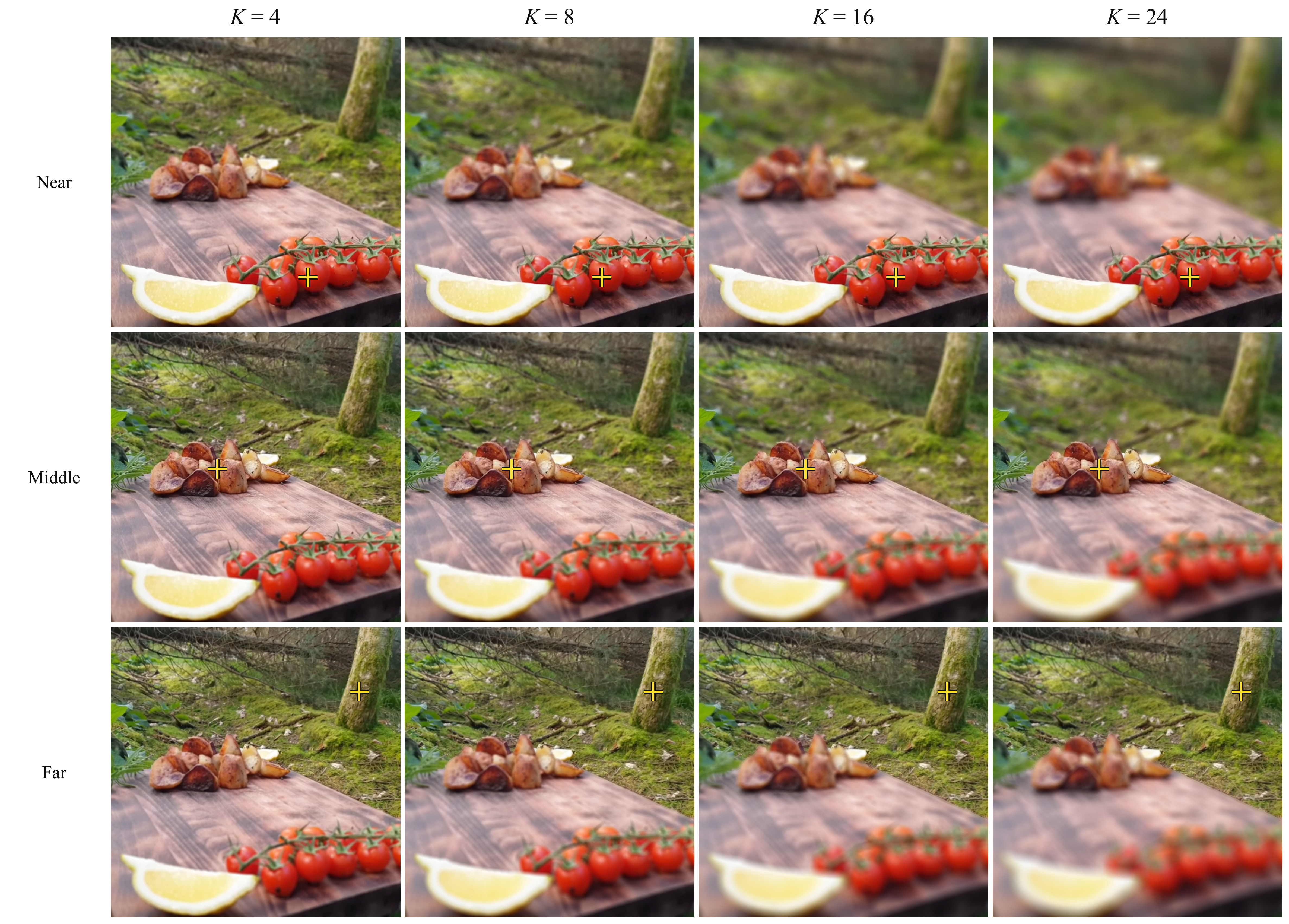}
    \caption{\textbf{Focus and blur strength control.}
    Rows select near, middle, or far focus locations, while columns use
    $K\in\{4,8,16,24\}$. All results show the same frame of the same
    source video. Yellow crosses indicate the target focus locations.}
    \label{fig:supp_focus_blur_control}
\end{figure}

\label{sec:supp_focus_blur_control}
Figure~\ref{fig:supp_focus_blur_control} demonstrates
RefocusCraft's control over target focus location and
blur strength on the same source video.
With $K$ fixed, changing the target focus location
relocates the relatively sharp region across near,
middle, and far scene depths, with corresponding
adjustments to defocus elsewhere.

With the focus location fixed, increasing $K$ from $4$
to $24$ progressively strengthens out-of-focus blur
while retaining comparatively clear structures near
the selected focus point.
These results demonstrate that RefocusCraft responds
effectively to both control conditions, enabling
flexible focus placement and blur strength adjustment
while preserving scene content.

\subsection{More Visual Comparisons}
\begin{figure}[!htb]
    \centering
    \includegraphics[width=.90\linewidth]{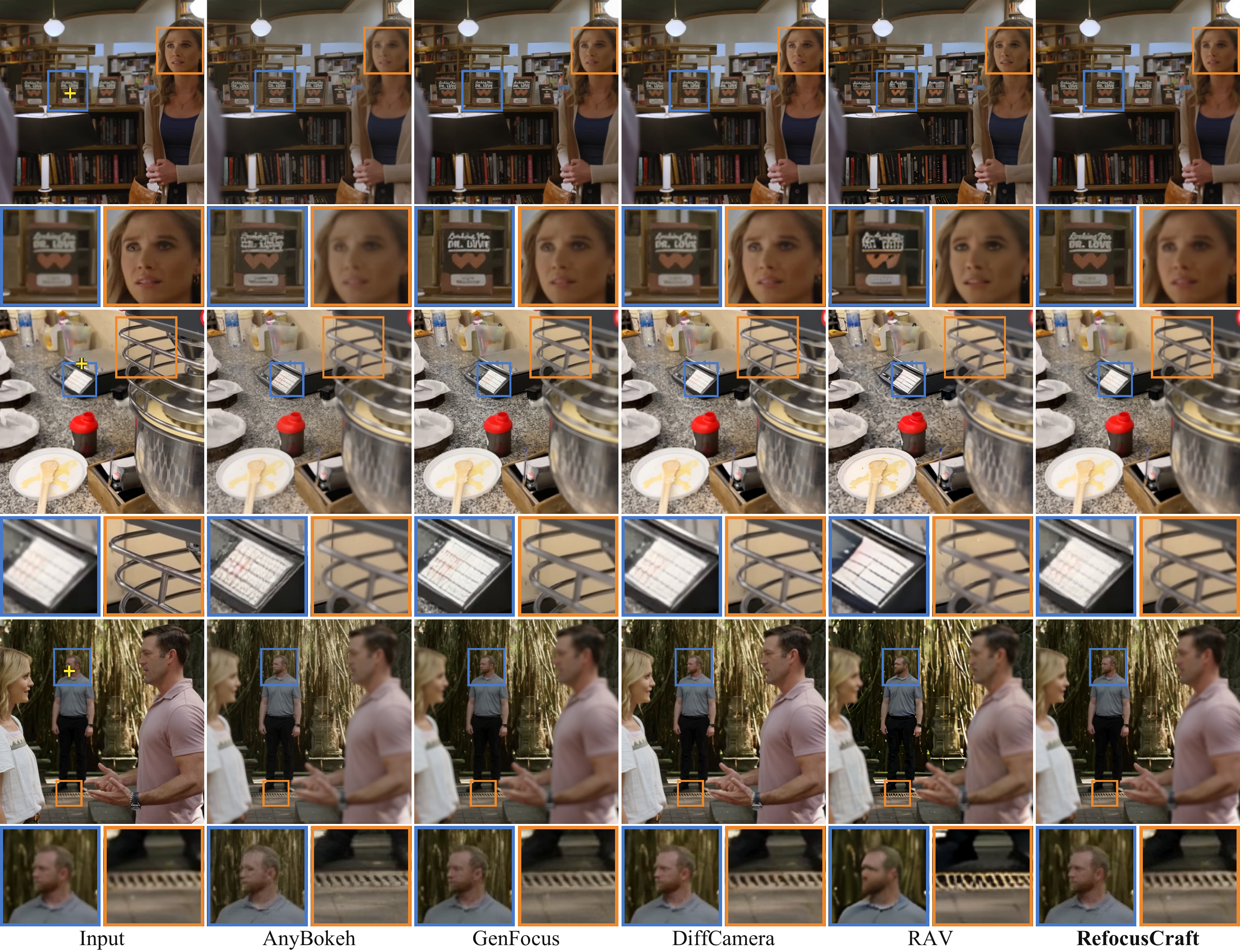}
    \caption{\textbf{Additional qualitative comparisons on real-world videos.}
    Yellow crosses mark target focus points; colored boxes indicate
    enlarged regions highlighting local details and refocusing effects.}
    \label{fig:supp_real_comparison}
\end{figure}

\label{sec:supp_visual_details}
Figure~\ref{fig:supp_real_comparison} presents additional
comparisons on real-world videos. RefocusCraft enhances
details in the target regions while retaining structures
closer to the source. In the bookshop and kitchen examples,
GenFocus and RAV introduce changes in the book-cover
lettering or fine texture patterns. Our results better
preserve these structures, particularly the grid pattern
in the enlarged kitchen region, while softening foreground
metal structures without pronounced shape changes.

The human-subject examples further highlight selective
defocus control and appearance preservation.
DiffCamera leaves the foreground people relatively sharp,
whereas RefocusCraft softens them while emphasizing the
selected background regions. In the outdoor example,
RefocusCraft also preserves the background subject's
facial and clothing appearance more faithfully than RAV.
These comparisons demonstrate the ability of RefocusCraft
to coordinate local detail enhancement and blur synthesis
while preserving scene content in real-world video
refocusing.

\section{Limitations}
\label{sec:supp_limitations}
\begin{figure}[!htb]
    \centering
    \includegraphics[width=.95\linewidth]{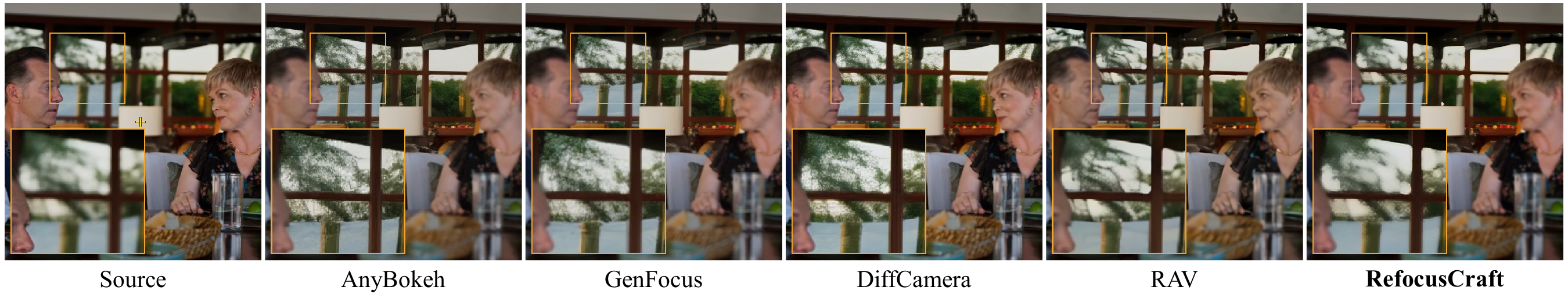}
    \caption{\textbf{A challenging real-world refocusing example.}
    The source and results from all compared methods show the same frame.
    The yellow cross marks the target focus point in the source.
    Orange boxes indicate a shared window region, enlarged in the insets.}
    \label{fig:supp_limitations_comparison}
\end{figure}

Like existing refocusing methods, RefocusCraft has difficulty
recovering reliable details from severely defocused source
videos. Strong defocus suppresses high-frequency information,
making the underlying textures ambiguous.
In Fig.~\ref{fig:supp_limitations_comparison}, the compared
methods produce noticeably different background textures,
while our result still retains residual blur in the enlarged
window region. Source feature reuse is constrained by the
information available in the input and cannot guarantee
faithful reconstruction when the original details are
heavily degraded.

\Needspace{10\baselineskip}
Our method also depends on source CoC estimation.
We currently apply the image-based estimator from
AnyBokeh~\cite{hou2026anybokeh} independently to each frame.
Without explicit temporal constraints, estimation errors
and frame-to-frame fluctuations may introduce inaccurate
or temporally inconsistent defocus guidance, affecting
the refocusing results. Repeated per-frame estimation
also introduces additional preprocessing overhead and
can become a bottleneck in video processing.
The reported refocusing runtime assumes precomputed
conditions and excludes this overhead.
Developing an efficient, temporally consistent video
CoC estimator is therefore an important direction
for future work.

\FloatBarrier

\end{document}